\documentclass{article} 
\usepackage{iclr2026_conference,times}

\usepackage{amsmath,amsfonts,bm}

\def\eqref#1{equation~\ref{#1}}

\def\1{\bm{1}}

\DeclareMathAlphabet{\mathsfit}{\encodingdefault}{\sfdefault}{m}{sl}
\SetMathAlphabet{\mathsfit}{bold}{\encodingdefault}{\sfdefault}{bx}{n}

\usepackage{xspace}
\usepackage{hyperref}
\usepackage{url}
\usepackage{mathtools}
\usepackage{float}
\usepackage{subfig}
\usepackage{graphicx}
\usepackage{textcomp}
\usepackage{xcolor}
\usepackage{listings}
\usepackage{xspace}
\usepackage{microtype} 
\usepackage{footmisc}  
\usepackage{xspace}
\usepackage{booktabs}
\usepackage{array}
\usepackage[ruled,linesnumbered]{algorithm2e}
\usepackage{wrapfig}
\usepackage{makecell}   
\usepackage{arydshln}   
\usepackage{diagbox}    
\usepackage{hyperref}
\usepackage{subcaption}

\usepackage{multirow}   
\usepackage{pifont}     

\hypersetup{hidelinks}

\NewDocumentCommand{\var}{O{s} m O{}}{%
  \ensuremath{#1_{#2}^{#3}}
}
\usepackage{siunitx}
\usepackage{colortbl}

\newcommand{\commentout}[1]{}

\definecolor{light-gray}{gray}{0.80}

\newcommand\fref{Fig.~\ref}

\newcommand\sref{Section~\ref}

\usepackage{amsthm}

\newtheorem{mytheorem}{Theorem}
\newtheorem{assumption}[mytheorem]{Assumption}
\newtheorem{lemma}[mytheorem]{Lemma}
\newtheorem{remk}[mytheorem]{Remark}

\newcommand{\hide}[1]{}

\renewcommand{\emph}[1]{\textit{#1}}

\newcommand{\sys}{AutoSP\xspace}

\title{\sys: Unlocking Long-Context LLM Training Via Compiler-Based Sequence Parallelism}

\iclrfinalcopy

\author{Ahan Gupta\thanks{Equal contribution.} , Zhihao Wang$^{*}$, Neel Dani \\
SSAIL Lab, University of Illinois Urbana-Champaign \\
Champaign, IL, USA \\
\texttt{\{ag82, zhihaow6, neeld2\}@illinois.edu} \\
\And
Masahiro Tanaka \\
Anyscale \\
San Francisco, CA, USA \\
\texttt{mtanaka@anyscale.com} \\
\And
Olatunji Ruwase \\
Snowflake \\
San Mateo, CA, USA \\
\texttt{tunji.ruwase@snowflake.com}
\And
Minjia Zhang \\
SSAIL Lab, University of Illinois Urbana-Champaign\\
Champaign, IL, USA \\
\texttt{minjiaz@illinois.edu} 
}

\begin{document}

\maketitle

\begin{abstract}
Large-language-models (LLMs) demonstrate enormous utility in long-context tasks which require processing prompts that consist of tens to hundreds of thousands of tokens. However, existing LLM training libraries do not provide easy to use abstractions to optimize for long-context training, instead focusing on optimizations for models with large parameter counts through ZeRO-3/FSDP, Tensor and Pipeline parallelism. This forces users to rewrite LLM training libraries to incorporate compositions of various complex long-context optimizations, such as sequence-parallelism, to training pipelines; a process that requires in-depth expertise, reducing developer productivity. To tackle these challenges, we introduce \sys: the first automated solution to automatically optimize LLM training for longer-contexts. \sys compiles models and applies a targeted set of optimizations: automated sequence parallelism, and long-context aware activation-checkpointing, to drastically enhance LLM trainability at negligible cost to throughput. Our evaluation demonstrates \sys's capability on both NVIDIA and AMD hardware, increasing training contexts by upto 2.7$\times$ and 2.5$\times$ respectively over competitive hand-written baseline at negligible cost to runtime performance.
\end{abstract}

\section{Introduction}
\label{sec:introduction}

Large Language Models (LLMs) are increasingly being trained with long-context data for scenarios such as document understanding~\citep{mdoc, simple-document-understanding, dude}, multi-step reasoning~\citep{gsm8k, smore}, and extended multi-turn dialogue generation~\citep{llama-2, mt-bench}. These use cases often contain input sequences ranging from tens to hundreds of thousands of tokens, creating massive activation memory demands and pushing the memory and system limits of GPU clusters. 

To circumvent out-of-memory errors, researchers have explored Sequence Parallelism (SP), a key enabler for long-context training. State-of-the-art SP strategies such as DeepSpeed-Ulysses~\citep{deepspeed-ulysses} and RingAttention~\citep{ring-attention} distribute the sequence dimension of activations across devices and allow the training engine to leverage aggregated GPU memory to train longer contexts with increasing device counts.

Despite effectively enabling long-context training, existing SP are implemented in eager mode and tightly coupled to specialized training frameworks such as DeepSpeed~\citep{deepspeed-zero} and Megatron-LM~\citep{megatron-lm}. Integrating SP to new training pipelines typically requires invasive code refactoring, which makes it difficult to apply across diverse model architectures and hardware platforms. Developers must manually insert communication collectives (e.g., \texttt{all2all}) between operators that require the full input sequence (such as attention), manage activation layouts across devices, and ensure correctness in both forward and backward passes. These manual efforts reduce scientists' productivity and limit portability. 

To improve productivity, researchers have begun to lift several complex distributed training strategies such as ZeRO-3/FSDP~\citep{deepspeed-zero} into SoTA deep-learning compilers, e.g. PyTorch-2.0~\citep{pytorch-2}. Examples include: SimpleFSDP~\citep{simple-fsdp} \& DeepCompile~\citep{deep-compile}, which implement ZeRO-3/FSDP as a series of compiler passes in the PyTorch-2.0 ecosystem. However, each of these techniques focuses on how to increase model parameter counts, uncovering different ways to shard model parameters rather than explicitly optimize for long-context training. While these efforts successfully lift data and model parallelism into compiler abstractions, they do not address parallelism strategies tailored to long-context training. This raises the question: can SP also be lifted into a deep-learning compilation stack to enable automated sequence parallelism?

\begin{wrapfigure}[16]{r}{0.50\textwidth}
\vspace{-15pt}
    \input{listings/teaser}
\end{wrapfigure}

In this work, we focus on lifting SP into PyTorch-2.0's compiler ecosystem as a compiler-based, PyTorch-native implementation. However, achieving this introduces several challenges. First, PyTorch-2.0's compilation pipeline includes multiple intermediate-representations (IRs) such as Torch-IR, Aten-IR, and Inductor-IR, to name a few. Each IR operates at a different abstraction level and encodes the input program at varying levels of granularity. This makes identifying an appropriate IR to conduct program-analysis, so as to recover the necessary information to apply semantically-preserving rewrites that transform single-GPU code into distributed sequence-parallel execution, a major challenge. A fine-grained abstraction will make program-analysis (to uncover important attributes about the input model) increasingly non-trivial whilst a coarse-grained abstraction will make semantic-rewrites (to insert communication collectives, transform buffer sizes and recompute manually indexed tensors) infeasible. Second, inferring sequence-dependent tensor shapes for resizing intermediate buffers is non-trivial to do within a compiler. (1) The lowering process inserts intermediate data-movement operators (such as transpositions), frequently changing the sequential axis of buffers. (2) SP strategies require resizing the sequential axis of only certain buffers (such as token and position id buffers), leaving other buffers (such as attention masks) untouched. Consequently, disambiguating which token's sequential axis requires resizing is challenging. Third, lifting SP into the PyTorch-2.0 compilation stack has consequences to other optimization passes that PyTorch-2.0 natively supports, notably: activation-checkpointing (AC). AC also enables memory savings for training by discarding activations in the forward pass and rematerializing them in the backwards pass to compute gradients. However, naively rematerializing activations together with SP triggers extraneous communication in the backwards pass, adversarially impacting runtime performance.

To tackle these challenges, we introduce \sys, the first compiler-based, PyTorch-native implementation of sequence parallelism. \sys introduces two key components: (1) a sequence-parallel transformation pass that automatically inserts communication collectives and reshapes activations, and (2) a sequence-aware activation checkpointing pass that exploits the compute-memory characteristics of long-context training. 
With just a few lines of code (as shown in \ref{teaser-code}), scientists can compile standard PyTorch models into distributed long-context training pipelines that scale input lengths without manual engineering. 
Our evaluation demonstrates that \sys significantly improves trainability at negligible loss to training speed on diverse hardware backends (NVIDIA and AMD GPUs), enabling training with up to 2.7$\times$ longer input context lengths compared to hand-written SP implementations such as DeepSpeed-Ulysses and RingAttention.

\section{Background}
\label{sec:background}

\begin{wrapfigure}{r}{0.5\textwidth}
    \centering
    \includegraphics[width=0.48\textwidth]{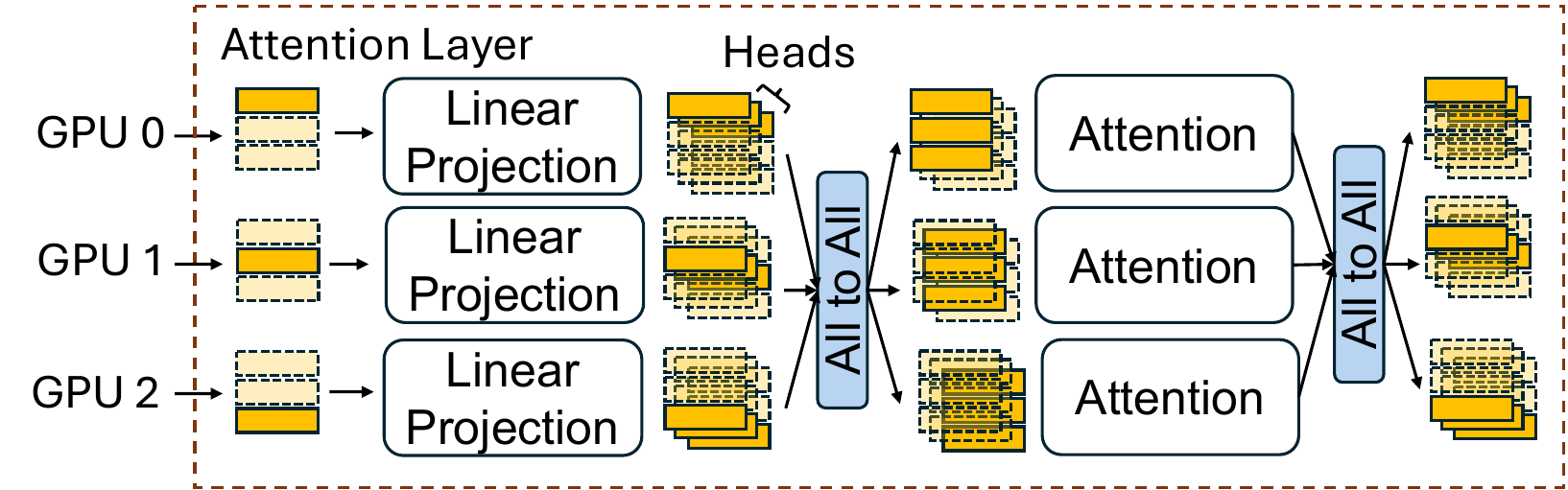}
    \caption{DeepSpeed-Ulysses with 3-SP groups. \texttt{alltoall} operators toggle the layout of activations at attention-layer boundaries.  
    Linear-projections operate on the partial sequence length, while attention-layers operate on a subset of the heads. 
    }
    \label{fig:ds-ulysses}
\end{wrapfigure}

\paragraph{Sequence parallel training.} Sequence parallelism (SP) is a key enabler for long-context training. These strategies enable scaling input sequence lengths with increasing GPU resources by sharding input tensors and activations across the sequence dimension. Communication collectives are inserted in the forward and backward pass as necessary to correctly shuffle tokens to the desired device. A popular SP strategy, and the focus of this work, is DeepSpeed-Ulysses (Ulysses)~\citep{deepspeed-ulysses}. We illustrate how Ulysses operates with 3-SP groups in ~\fref{fig:ds-ulysses}. 
First, tokens are sharded across the sequence dimension, with different devices operating on different parts of the input sequence. Next, linear projections form multiple \texttt{Q/K/V} heads. Since linear projections operate pointwise across the sequential dimension, each device directly operates on its own tokens with no additional communication. However, since attention requires the entire input-context, an all-to-all reshuffles tokens, re-sharding the activations across the head-dimension. Each device locally computes attention on its respective head(s) after which another all-to-all reshuffles the tokens back to their original input sizes, re-sharding across the sequence dimension.

\begin{wrapfigure}[24]{r}{0.40\textwidth}
\vspace{-15pt}
    \centering
    \includegraphics[width=\linewidth]{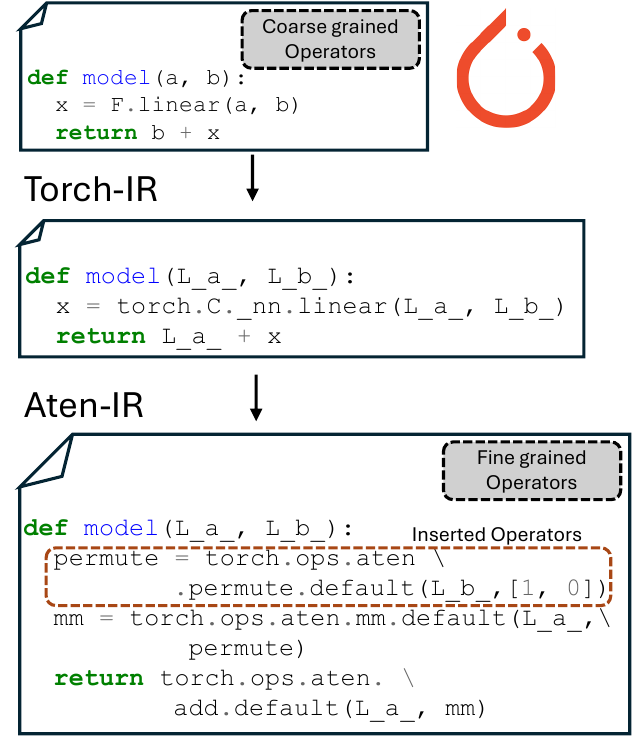}
    \caption{A sample neural network compiled using PyTorch-2.0. We illustrate the lowering to Torch-IR and Aten-IR that occur within Dynamo and the extra operators inserted during the lowering process.}
    \label{fig:torch-lowering}
\end{wrapfigure}

\paragraph{PyTorch-2.0 compiler.} PyTorch-2.0~\citep{pytorch-2} is a just-in-time deep-learning compiler that targets training workloads. It comprises of two components: dynamo and inductor each with a series of \textit{compiler passes} that progressively lowers and optimizes code. An example code-snippet of a neural-network and its progressive lowering through different IRs, is shown in~\fref{fig:torch-lowering}.

\textit{Dynamo.} The input to dynamo is a model comprising of PyTorch \& python operators. Dynamo then executes the function and records a \textit{trace}, represented as its intermediate representation: a computation graph. Each node in the computation-graph comprises of Torch-IR statements, which loosely correspond to statements in the original input program. Next, AOTAutograd lowers each Torch-IR statement to Aten-IR statements, which consist of finer-grained operators. Aten-IR statements do not consist of higher-level abstractions such as linear or attention layers, but instead consist of (batch) matrix-multiplication, convolution, and data-movement operators. Each Torch-IR statement is accordingly lowered to its corresponding set of (multiple) Aten-IR statement(s), forming an \textit{FX-graph}. For example, in~\fref{fig:torch-lowering}, we observe that the \texttt{linear} operator in Torch-IR is lowered to two operators in Aten-IR: \texttt{permute} and \texttt{mm}. At this stage, a variety of compiler-passes to optimize the FX-graph are applied, notably automated acitvation-checkpointing (AC). The AC compiler-pass~\citep{pytorch-ac} is responsible for selecting which tensors to rematerialize in the backwards pass without incurring performance penalties. It reduces this problem to a network-flow construction whose min-cut determines the tensors to rematerialize. We describe it in detail in~\sref{subsec:auto-ac}.

\textit{Inductor.} Finally, inductor consumes the output Aten-IR FX-graph and lowers it to a custom define-by-run IR, subsequently code-generating necessary kernels specialized to the backend microarchitecture.

\section{\sys}
\label{sec:system}

\begin{figure*}[!t]
    \centering
    \includegraphics[width=\linewidth]{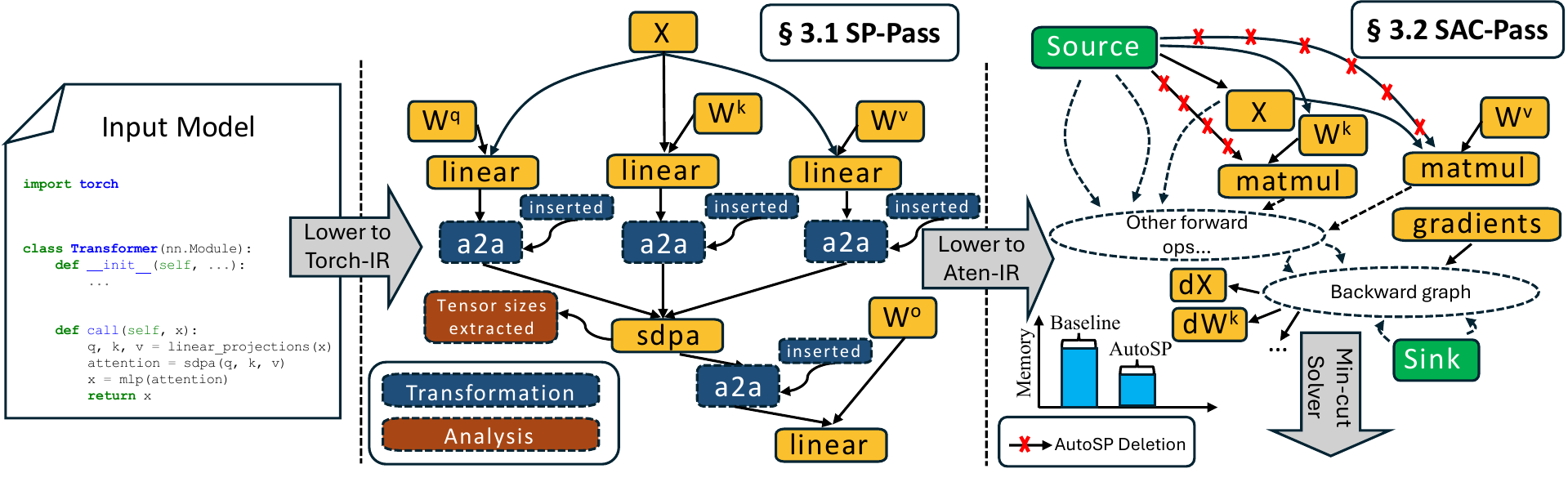}
    \caption{An overview of \sys. \sys enables an automated approach to scale input context lengths for long-context training through a targeted set of compiler optimization passes: an automated sequence-parallelism pass (\sref{subsec:auto-sp}), and a SP-aware long-context AC-pass: SAC (\sref{subsec:auto-ac}).}
    \label{fig:autosp-overview}
\end{figure*}

\sys lifts SP parallelism as a compiler-pass into the PyTorch-2.0 compiler stack to optimize long-context training. ~\fref{fig:autosp-overview} 
is an overview of how \sys's compiler-passes interoperate with the PyTorch-2.0 compilation stack to optimize LLM training code. ~\sref{subsec:auto-sp} describes how we enable automated sequence parallelism as a compiler pass, and~\sref{subsec:auto-ac} describes our long-context aware activation checkpointing strategy. 

\subsection{Automated Sequence Parallelism} 
\label{subsec:auto-sp}

\textbf{Challenges.} To implement automated sequence parallelism, we must select (a single) PyTorch IR(s) to analyze and transform accordingly. However, recovering the necessary model information through program analysis and subsequently transforming the model through semantically-preserving rewrites is non-trivial for three reasons. (1) Each inserted communication collective must have token-buffers instantiated to a particular size parameterized by the model-dimension, batch-size and sequence length. However, these parameters are not explicitly represented in any IR. (2) Existing intermediate buffers need to be resized to different shapes depending on the SP group-size as well as their placement within the neural network. For example, in DeepSpeed-Ulysses, buffers within the attention layer should be resized to operate on the full-sequence but a subset of all the heads, whilst buffers within MLP layers operate on the full model-dimension but on a partial sequence-length. (3) Certain operations require manual indexing for correctness, e.g. indexing the causal mask to appropriately apply to attention matrices, and need to be automatically recomputed.  

To tackle these challenges, \sys's SP-pass accordingly analyzes computational structures to extract the pertinent information required for transforming single-GPU code to distributed sequence-parallel code. We next describe how it operates.

\subsubsection{Analysis and Transformations} Our pass operates in two stages. First, we analyze tensor sizes to gather information about the batch, sequence and hidden dimension that parameterize the model. Next, we transform the IR by: (1) inserting communication collectives at appropriate places in the network with appropriately sized buffers to store their output, (2) adjusting the sizes of existing buffers within the graph to account for sequence sharding. (3) Recompute any manually indexed tensors appropriately. All our analysis and transformations operate on Torch-IR.

\textbf{Why AutoSP analyzes and transforms Torch-IR?} Our SP-pass operates on Torch-IR as the analysis and transformations are significantly more challenging to accomplish on Aten-IR for three reasons. (1) Torch-IR more closely resembles the neural-network programmed by the user (see~\sref{sec:background}) with operators such as linear and attention-layers, making it easier to identify which parts of the graph belong to which layer for appropriate tensor resizing. On the other hand, equivalent operators in Aten-IR are represented as a series of finer-grained operators, such as mat-muls and permutations, making it challenging to reason about which operators belong to linear-projections and attention-operators respectively. (2) The lowering process from Torch-IR to Aten-IR inserts various data-layout transformations such as reshapes and permutes, obscuring information as to which dimension corresponds to the sequence, batch, and hidden sizes of a tensor, making it challenging to appropriately resize the correct tensor dimension. (3) Torch-IR operates on only the forward-pass resulting in our transformations operating on a simpler computation-graph, merely requiring each new added node to have a registered conjugate gradient operator. On the other hand, Aten-IR operates on both the forward and backwards pass and requires more complex transformations to the computation-graph.

\begin{wrapfigure}[26]{r}{0.50\textwidth}
    \vspace{-14pt}
    \input{listings/transformation_pass}
\end{wrapfigure}

\textbf{Program analysis to uncover training parameters.} To correctly instantiate token-buffers for communication collectives, the correct batch, sequence and model-dimensions need to be extracted. Fortunately, we can analyze the input nodes of the entire computation graph to extract the necessary information. Since the input to the computation-graph is the data-loaded after preprocessing, it is \textit{guaranteed} to resemble a particular shape depending on the problem domain. For example, in natural-language tasks, the data will be a \texttt{[batch, seq\_length]}-sized tensor.
Next, to acquire the model-dimension we traverse the graph until we encounter an attention operator and inspect its output ND tensor whose last two dimensions are: \texttt{[num\_heads, head\_dim]}-sized. The product of the outer two dimensions is the model-dimension.

\textbf{Program transformation.} After acquiring the batch, sequence and model-dimensions, we have the necessary information to transform the computation graph from single-GPU to distributed sequence-parallel code. Listing~\ref{transformation-code} illustrates how we transform the existing computation-graph, \texttt{mod}, comprising of Torch-IR statements. We traverse through the graph, and for each node: (1) Check if it belongs to the \texttt{RESIZE\_BUFS} set, and accordingly resize its buffers depending on its placement in the attention or linear-projection/MLP-layers. (2) Check if it belongs in the \texttt{INDEX\_OPS} set, and accordingly resize its tensor indexing. (3) Check if it is the first/last attention-op and accordingly instantiate communication buffers and insert the necessary \texttt{alltoall} before/after the attention-layer. We manually curate the \texttt{RESIZE\_BUFS}, \texttt{INDEX\_OPS}, and \texttt{ATTN\_OPS} sets by analyzing dynamo FX-graphs of compiled hand-written transformer implementations.

\subsection{Sequence-Parallel aware Activation Checkpointing}
\label{subsec:auto-ac}

\textbf{Challenges.} In addition to the SP-pass, 
activation checkpointing (AC) is an important memory reducing optimization that enables longer context training. \texttt{torch.compile} provides an automated AC-pass~\citep{pytorch-ac} that operates on Aten-IR to compose with arbitrary neural networks. 
However, naively composing its AC-pass with \sys's SP-pass within the compiler-stack leads to sub-optimal performance for long-context training. We briefly explain how PyTorch-2.0's automated AC-pass functions, and why it is insufficient.

\begin{wrapfigure}{r}{0.45\textwidth}
    \vspace{-13pt}
    \centering
    \includegraphics[width=\linewidth]{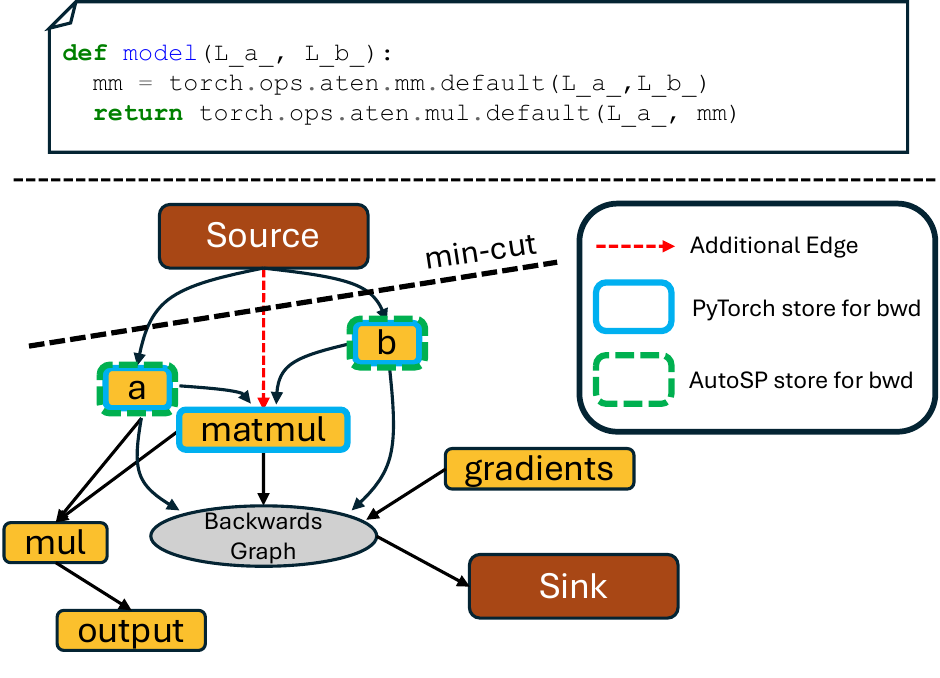}
    \caption{A comparison of PyTorch-2.0 vs. \sys's AC-passes on a sample code-snippet. The red line is the additional edge from source to node in PyTorch-2.0's AC-pass to enforce no rematerialization of compute heavy operators. \sys's AC-pass, in removing this constraint, reduces memory-consumption at negligible cost to throughput.}
    \label{fig:ac-pass}
\end{wrapfigure}

\textbf{PyTorch-2.0's AC-pass.} PyTorch's AC-pass operates on Aten-IR, within Dynamo.
Its primary function is to reduce memory consumption of model training without incurring performance penalties. The optimization reduces the problem to a network-flow construction whose min-cut determines the tensors to rematerialize. We give an example code-snippet and its equivalent network-flow construction in~\fref{fig:ac-pass}. The input to the optimization is an FX-graph comprising of Aten-IR statements. First, a \textit{joint-graph} of the forward and backwards graph is constructed. Next, the source node is connected to all the input tensors of the graph, and all the nodes reachable from the incoming gradients are connected to the sink. Then, capacities, representing costs, are assigned to each node. A node's capacity is determined by a heuristic function comprising of various characteristics such as the output activation memory produced. Finally, the problem is converted from a flow on nodes to a flow on edges. Each edge's capacity is set to \texttt{inf} and each node is split into two: an \texttt{\_in} and \texttt{\_out} node. Incoming edges to each node are connected to its respective \texttt{\_in} node and outgoing edges are connected to its respective \texttt{\_out} node. An edge with the original node's capacity connects \texttt{\_in} to \texttt{\_out}. A min-cut on this graph will cut only finite-capacity edges from a node's \texttt{\_in} to \texttt{\_out} (highlighted by green circles in~\fref{fig:ac-pass}); only nodes on this cut are stored. Intuitively, the min-cut identifies the smallest cost set of activations to preserve to compute the necessary dependencies in the backwards pass. PyTorch-2.0 additionally enforces constraints on which nodes can be rematerialized, resulting in its conservative nature which we explain next.

\textbf{Why PyTorch-2.0's AC solution is insufficient.} PyTorch-2.0's AC-pass is effective in identifying which activations to store without compromising runtime performance in polynomial time. However, it makes the conservative decision to disallow rematerialization of many \textit{seemingly} compute-intensive operators such as: mat-muls, and scaled mat-muls to name a few, based purely on operator type, without considering runtime cost in long-context settings. This is enforced by connecting each compute-intensive node's \texttt{\_in} to the source with infinite capacity (indicated by the additional dotted red line in~\fref{fig:ac-pass}), enforcing that node, or some downstream value of it, to belong to the min-cut. However, in long-context training, we observe that certain compute-intensive operators take a small fraction of overall compute and can accordingly be rematerialized without incurring performance penalties. \sys's AC strategy exploits this. We explain these observations next.

\textbf{Observations.} We analyze the structure of compute operations in modern LLMs to identify regions that can be appropriately rematerialized. For a transformer with: batch-size \texttt{b}, sequence-length \texttt{s}, number of heads \texttt{h}, head dimension \texttt{d}, and MLP hidden dimension \texttt{d$_\text{ffn}$}, we have that: \begin{align*}
    \texttt{2bhs$^\text{2}$d} &\text{    Attention FLOPS} \\
    \texttt{8bhsd$^\text{2}$} &\text{    Linear-projection FLOPS} \\
    \texttt{4bhsd$_\text{ffn}$d} &\text{    MLP FLOPS} \\
\end{align*} 
\par\vspace{-2em}
When training on long-contexts, we have that \texttt{s>>d,h,d$_\text{ffn}$}, which results in the following fraction of FLOPs linear-projection and MLP layers take over all the compute operations: \begin{equation}
    \frac{\texttt{8bhsd$^\text{2}$ + 4bhsd$_\text{ffn}$d}}{\texttt{2bhs$^\text{2}$d} + \texttt{8bhsd$^\text{2}$ + 4bhsd$_\text{ffn}$d}} \approx O\bigg(\frac{\texttt{1}}{\texttt{s}}\bigg) \quad \text{  as  } \texttt{s} \to \infty \label{eqn:fraction}
\end{equation} Indicating that the fraction of compute operations for linear-projection and MLP matrix-multiplications decreases as a function of input-sequence length. This observation underpins \sys's automated AC strategy.

\textbf{\sys's AC strategy.} \sys exploits equation~\ref{eqn:fraction}, building upon PyTorch-2's automated AC strategy. However, instead of conservatively banning \textit{every} compute-intensive operator, we permit configurations where (batch) matrix-multiplications and other compute-intensive operators outside of the attention layer are rematerialized.
We achieve this by iterating over the joint-graph and removing any additional edges from the source to compute-heavy operators, resulting in only inputs to the graph (tensors \texttt{a} and \texttt{b} in~\fref{fig:ac-pass}) connecting to the source. We then dispatch this mutated joint-graph to PyTorch-2.0's AC strategy. This change enables traning on significantly longer context lengths at negligible cost to training throughput. 

\section{Evaluation}
\label{sec:evaluation}

We evaluate \sys with a comprehensive set of experiments. We demonstrate its effectiveness in enhancing trainibility of various models and sizes in~\sref{subsec:main-results}, and detailed breakdowns of the impact of each component in~\sref{subsec:analysis}.


\textbf{Setup.} We evaluate \sys and all the baselines on NVIDIA GH200-96GB \& A100-80GB and AMD MI250-64GB hardware. All experiments use PyTorch-2.7 with CUDA 12.8 (on NVIDIA GPUs), and ROCm 6.4 (on AMD GPUs). To implement \sys, we lift the DeepSpeed-Ulysses SP scheme into PyTorch-2.0's compilation stack and integrate all our compiler optimizations into the DeepSpeed project, due to its popularity in training large scale LLMs.

\textbf{Baselines.} We compare \sys to both compiler-optimized distributed training solutions and hand-optimized SP solutions to demonstrate the memory and compute efficiency of our approach. Specifically, we compare against ZeRO-3 (FSDP)~\cite{deepspeed-zero} optimized through \texttt{torch.compile()} in PyTorch-2.0, and hand-written DeepSpeed-Ulysses~\citep{deepspeed-ulysses} \& RingAttention implementations when compiled under PyTorch-2.0's inductor backend. We use DeepSpeed-Ulysses' code in the original DeepSpeed repository\footnote{https://github.com/deepspeedai/DeepSpeed/tree/master/deepspeed/sequence} and RingFlashAttention\footnote{https://github.com/zhuzilin/ring-flash-attention}, which are both highly optimized implementations of SP strategies. We evaluate all techniques on a range of model sizes: Llama-3.2 1B \& 3B, Llama-3.1 8B, and Llama-2 13B, covering models with either Grouped-Query-Attention (GQA) or Full-Attention. 

\subsection{Main Results}
\label{subsec:main-results}

In this section, we evaluate how \sys impacts model trainability: the maximum trainable sequence length prior to encountering OOM issues. 

\begin{figure}[!ht]
    \centering
    \begin{minipage}[t]{0.47\textwidth}
        \centering
        \includegraphics[width=\textwidth]{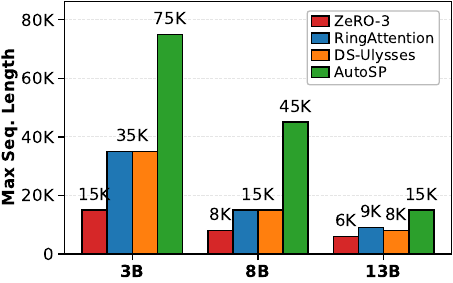}
        \caption{Maximum sequence length prior to OOM across various model sizes. \sys increases the trainability of all model sizes.}
        \label{fig:model-size-trainibility}
    \end{minipage}
    \hfill
    \begin{minipage}[t]{0.47\textwidth}
        \centering
        \includegraphics[width=\textwidth]{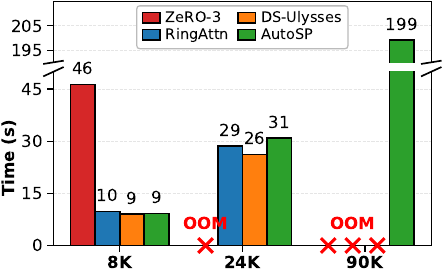}
        \caption{End-to-End training iteration times (over 10 iterations) for Llama-3.1 8B on 8 GPUs. \sys maintains similar per-training iteration time whilst significantly increasing trainability.}
    \label{fig:scalability}
    \end{minipage}
\end{figure}

\textbf{Trainability.} For different techniques, we measure the maximum sequence length trainable prior to OOM on 8 NVIDIA A100-80GB in~\fref{fig:model-size-trainibility}. For all models, we use ZeRO-1, setting the SP group-size to 2 and the DP group-size to 4 with the exception of the ZeRO-3 augmented with \texttt{torch.compile()} baseline. Compared to ZeRO-3 (FSDP), \sys enables training on upto 5$\times$, 5.6$\times$ and 2.5$\times$ longer input sequences for the 3B, 8B and 13B models, respectively. The trainability gains come from \sys's compiler-based SP-pass, an optimization that targets long-context training unlike the ZeRO-3 baseline, which instead targets models with large parameter counts. Moreover, \sys achieves significant trainability gains compared to both inductor compiled DS-Ulysses and RingAttention implementations. Compared to DS-Ulysses, \sys enables training on longer input sequences by upto 2.14$\times$, 3$\times$ and 1.88$\times$ for the 3B, 8B and 13B models, respectively. Compared to RingAttention, \sys enables training on upto 2.14$\times$, 3$\times$ and 1.6$\times$ for the 3B, 8B and 13B models, respectively. The additional gains over hand-written SP implementations come from the SP-aware AC-pass that exploits equation~\ref{eqn:fraction}, rematerializing compute-heavy operators (such as mat-muls) for low runtime costs but large memory gains. The trainability gains are especially pronounced for the 8B model compared to 3B as many of the compute-heavy operators, such as linear-projections and MLPs, produce activations parameterized by the model hidden-dimension. These need to be stored to compute gradients in the backwards pass and result in larger models producing significantly more activation memory due to compute-heavy operators. \sys, in rematerializing these large tensors, alleviates memory issues at negligible runtime cost. The trainability gains are less pronounced for the 13B model as optimizer states begin to consume a substantial portion of memory ($\sim$50\%). 

\begin{wrapfigure}[19]{R}{0.45\textwidth}
    \centering
    \includegraphics[width=0.47\textwidth]{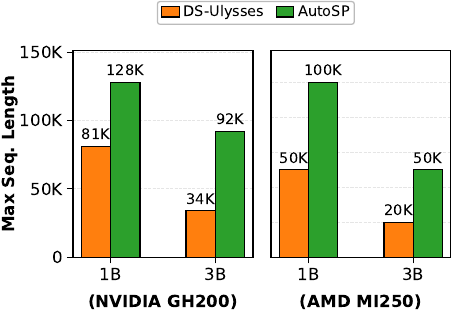}
        \caption{Comparing the max sequence length prior to OOM across different hardware. \sys enables training on longer sequences on NVIDIA superchips and AMD hardware.}
        \label{fig:hardware-portability}
\end{wrapfigure}

\textbf{Runtime Performance.} For different techniques, we identify the impact on training iteration times at various sequence lengths in~\fref{fig:scalability}. We focus on a Llama-3.1 8B model, using ZeRO-1 with a fixed DP-size of 2 and SP-size of 4, and evaluate on 8 GPUs, measuring time taken for 10 training iterations. We toggle the sequence-aware AC pass on for \sys only if OOM issues occur. Compared to the ZeRO-3 baseline compiled with \texttt{torch.compile()}, \sys reduces per-iteration times by 5$\times$ whilst increasing trainability by an order of magnitude. ZeRO-3 introduces more communication collectives over ZeRO-1 with SP, resulting in slower per-iteration times. Moreover, compared to RingAttention and DS-Ulysses, \sys increases trainability by 3.75$\times$ with negligible cost to runtime-performance. Finally, we observe that DS-Ulysses is faster than RingAttention due to RingAttention's p-step (where p is the SP size) communication latency which exchanges key/values in a ring-like pattern across SP-groups. Comparitively, DS-Ulysses introduces a single all-to-all to re-shard token-buffers. In modern clusters, intra-node all-to-alls are fast due to pairwise communication links between devices.

\subsection{Analysis}
\label{subsec:analysis}

\begin{wrapfigure}[27]{R}{0.45\textwidth}
\vspace{-13pt}
    \centering
    \includegraphics[width=0.47\textwidth]{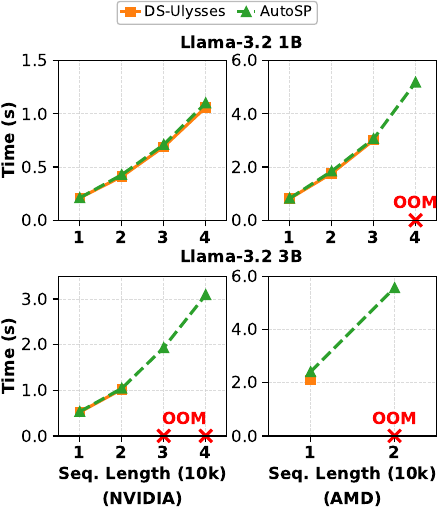}
    \caption{Average execution time of various Llama 3.2 model sizes at different sequence lengths on NVIDIA and AMD hardware. \sys matches the performance of hand-written baselines and supports longer sequence training.}
    \label{fig:runtime-performance}
\end{wrapfigure}
In this section, we run additional studies to demonstrate the effectiveness of \sys, as well as to ascertain the impact of each optimization on different components of LLM training. We compare \sys with DS-Ulysses only (rather than RingAttention) given it enables training on similarly sized context-lengths whilst being slightly faster. 

\textbf{Trainability on diverse hardware.} We run an additional trainibility study on different NVIDIA and AMD hardware, identifying the maximum sequence length before encountering an OOM in~\fref{fig:hardware-portability} to demonstrate \sys's portability. We focus on smaller, 1B and 3B models, running on either 2 GH200-96GB (NVIDIA) GPUs, or 2 AMD MI250-64GB GPUs. On NVIDIA hardware, \sys enables training on 1.58$\times$ (2.70$\times$) longer sequence lengths on the 1B (3B) models respectively. On AMD hardware, \sys enables training on 2$\times$ (2.5$\times$) longer input sequences on the 1B (3B) models, respectively. \sys consistently delivers significant trainability gains across diverse hardware, and model sizes through its targeted long-context optimizations.

\textbf{Runtime performance on diverse hardware.} For different techniques, we measure the per-iteration end-to-end training time (averaged across 100 training iterations) on NVIDIA (GH200-96GB) and AMD (MI250-64GB) hardware in~\fref{fig:runtime-performance}, demonstrating the performance-portability of our approach. We trigger the SP-aware AC-pass only to avoid OOM issues for \sys. On sequence lengths that all techniques can train on, \sys has the following speedups: 0.97$\times$ (1B) and 0.98$\times$ (3B) compared to the inductor baseline on NVIDIA hardware, respectively. On AMD hardware, \sys has the following speedups: 0.97$\times$ (1B) and 0.87$\times$ (3B) compared to the inductor baseline, respectively. We note two observations. Despite being a general and performance-portable compiler pass, \sys achieves 97\% of DS-Ulysses' hand-written baseline whilst providing an upto 2.7$\times$ trainability gain. Without \sys's targeted optimizations, training at long-contexts quickly becomes infeasible.  

\begin{wrapfigure}[17]{R}{0.45\textwidth}
    \vspace{-15pt}
    \centering
    \includegraphics[width=0.47\textwidth]{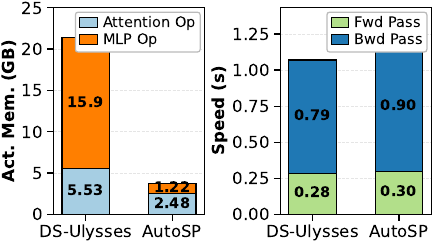}
    \caption{Breakdown of Attention and MLP operator memory consumption and forward iteration times. \sys reduces the activation memory of Attention and MLP operators with a marginal performance difference.}
    \label{fig:breakdown-analysis}
\end{wrapfigure}

\textbf{Breakdown analysis.} We breakdown the impact of \sys's optimizations in~\fref{fig:breakdown-analysis} on a NVIDIA GH200-96GB. We breakdown the activation memory produced by the attention and MLP operators as well as the per-iteration runtime of the forward and backward passes when training a Llama-3.2 1B model using a sequence length of 40k. Overall, \sys reduces memory consumption of the attention and MLP operators by 13.03$\times$ and 2.22$\times$, respectively. The marked impact on MLP operators arises due to the presence of many mat-muls, which \sys's AC-pass rematerializes to alleviate memory consumption at low runtime costs. On the other hand, \sys incurs a 1.14$\times$ cost runtime performance cost for the backward pass, whilst having similar forward pass times as a result of the extra rematerialized operators, which are recomputed in the backwards pass only.

\textbf{Ablation study.} Finally, we toggle different optimizations on/off to demonstrate their impact on trainability and speed in Table~\ref{tab:optimization_comparison}. The top row indicates the maximum trainable input context length before an OOM, and the bottom row indicates the average training iteration time at a fixed 40k input sequence length. Overall, \sys's optimizations result in a 1.58$\times$ trainability gain whilst achieving 89\% of DeepSpeed-Ulysses' per-iteration training time. Moreover, the incremental trainability gain of the AC-pass over the SP-pass is 1.66$\times$ with a mere 7\% decrease in runtime-performance. This significant trainability gain with little runtime-performance cost is due to equation~\ref{eqn:fraction}. At longer contexts, the attention-operator's FLOPs dominate runtime, enabling traditionally compute-heavy operators to be rematerialized with marginal performance costs and large memory gains. Note that the baseline is a highly hand-optimized SP implementation with several optimizations such as communication-computation overlapping using streams. Nevertheless, the SP-pass achieves 97\% of the baseline's performance as a general compiler-pass.

\begin{table}[h]
\centering
\caption{Various optimizations in \sys against a baseline on a Llama-3.1 1B. \sys supports training on significantly longer sequence lengths at minimal cost to performance.}
\label{tab:optimization_comparison}
\begin{tabular}{llcc}
\toprule
\multicolumn{2}{c}{\textbf{Method}} & \textbf{Max Tokens} & \textbf{Speed (s)} \\
\midrule
\multicolumn{2}{l}{DS-Ulysses} & 81,000 & \textbf{1.06} \\
\midrule
\multirow{2}{*}{AutoSP} & SP-Pass & 77,000 & 1.09 \\
 & SP \& AC-Pass & \textbf{128,000} & 1.19 \\
\bottomrule
\end{tabular}
\end{table}

\section{Related Work}
\label{sec:related-work}

\textbf{Parallel training strategies.} ZeRO-3/FSDP~\citep{deepspeed-zero}, Tensor~\cite{megatron-lm} \& Pipeline~\citep{gpipe} parallelism are training strategies that target models with large parameter counts, reducing per-device memory consumption of optimizer, model, activation and gradient states. Expert-parallelism~\citep{gshard, deepseek-v3, deepseekmoe} targets large sparse mixtures-of-experts~\cite{} which contain many intermediate expert MLPs. Though effective in enhancing the trainability of large-language model training, these parallel strategies do not explicitly target long-context training and are insufficient to scale input context lengths.

\textbf{Automated optimizations.} Deep-compile~\citep{deep-compile} provides an automated approach to implement ZeRO-3/FSDP using profile-guided optimization. Though effective, FSDP does not explicitly target long-context training. General-Single-Program-Multiple-Data (GSPMD)~\citep{gspmd}, is an automated parallelization strategy in XLA guided through user annotations, requiring some human effort. Lastly, deep-learning compilers such as: TVM~\citep{tvm}, Mirage~\citep{mirage}, and AITemplate~\citep{aitemplate}, focus on schedule rewrites for inference workloads only and do not consider inter-GPU parallelism, a key optimization for long-context training.

\textbf{Activation checkpointing.} Various works propose AC techniques~\citep{checkmate, dynamic-tensor-remat, sublinear-memory-cost}. They primarily consist of two approaches. (1) Search-based optimization (e.g. via integer-linear-programming), which may not scale up to today's LLM sizes (billion parameter models). (2) Static-policies (e.g. checkpoint chunks of $\sqrt{N}$ layers' activations), which may result in extraneous communication calls in the backward pass in the SP-setting. Comparatively, our SP-aware AC-pass exploits observations of compute \& memory properties of LLM training at long-contexts to alleviate memory consumption at negligible cost to runtime performance. Additionally, \sys's sequence-aware AC pass is implemented in PyTorch-2.0's compiler stack, requiring no human intervention. In contrast, existing AC strategies often require invasive code-rewrites.


\section{Conclusion}

In this paper, we present \sys, the first compiler-based, PyTorch-native solution for training large-language-models at long-contexts. Through a combination of automated sequence-parallelism (SP) and a sequence-aware AC strategy, \sys achieves significant sequence length extensions at negligible cost to training throughput. Our results demonstrate that compiler-driven, PyTorch-native automation provides a practical and portable foundation for long-context model training.

\section*{Reproducibility Statement}

We have made several efforts to ensure the reproducibility of our work. We have documented the critical components of our implementation in~\sref{sec:system} and our evaluation in~\sref{sec:evaluation} additionally documents our setup with a description of our hardware and software versions. We have ensured that each result is consistent with good benchmarking practices, including taking the average over multiple runs. Moreover, all our code and benchmarks will be made publicly available to further enhance reproducibility. 

%
%
%
%
%

\section*{Acknowledgments}

We sincerely appreciate the anonymous reviewers. Their insightful feedback helps significantly improve the quality of the paper. This research was supported by the National Science Foundation (NSF) under Grant No. 2441601. The work utilized the Delta and DeltaAI system at the National Center for Supercomputing Applications (NCSA) and Jetstream2 at Indiana University through allocation CIS240055 from the Advanced Cyberinfrastructure Coordination Ecosystem: Services \& Support (ACCESS) program, which is supported by National Science Foundation grants \#2138259, \#2138286, \#2138307, \#2137603, and \#2138296. The Delta advanced computing resource is a collaborative effort between the University of Illinois Urbana-Champaign and NCSA, supported by the NSF (award OAC 2005572) and the State of Illinois. UIUC SSAIL Lab is supported by research funding and gift from IBM, Google, Amazon, and AMD.

\bibliography{iclr2026_conference}

@misc{deepspeed-zero,
      title={ZeRO: Memory Optimizations Toward Training Trillion Parameter Models}, 
      author={Samyam Rajbhandari and Jeff Rasley and Olatunji Ruwase and Yuxiong He},
      year={2020},
      eprint={1910.02054},
      archivePrefix={arXiv},
      primaryClass={cs.LG},
      url={https://arxiv.org/abs/1910.02054}, 
}

@misc{megatron-lm,
      title={Megatron-LM: Training Multi-Billion Parameter Language Models Using Model Parallelism}, 
      author={Mohammad Shoeybi and Mostofa Patwary and Raul Puri and Patrick LeGresley and Jared Casper and Bryan Catanzaro},
      year={2020},
      eprint={1909.08053},
      archivePrefix={arXiv},
      primaryClass={cs.CL},
      url={https://arxiv.org/abs/1909.08053}, 
}

@inproceedings{deepspeed-ulysses,
author = {Jacobs, Sam Ade and Tanaka, Masahiro and Zhang, Chengming and Zhang, Minjia and Aminadabi, Reza Yazdani and Song, Shuaiwen Leon and Rajbhandari, Samyam and He, Yuxiong},
title = {System Optimizations for Enabling Training of Extreme Long Sequence Transformer Models},
year = {2024},
isbn = {9798400706684},
publisher = {Association for Computing Machinery},
address = {New York, NY, USA},
url = {https://doi.org/10.1145/3662158.3662806},
doi = {10.1145/3662158.3662806},
booktitle = {Proceedings of the 43rd ACM Symposium on Principles of Distributed Computing},
pages = {121–130},
numpages = {10},
location = {Nantes, France},
series = {PODC '24}
}

@misc{ring-attention,
      title={Ring Attention with Blockwise Transformers for Near-Infinite Context}, 
      author={Hao Liu and Matei Zaharia and Pieter Abbeel},
      year={2023},
      eprint={2310.01889},
      archivePrefix={arXiv},
      primaryClass={cs.CL},
      url={https://arxiv.org/abs/2310.01889}, 
}

@misc{simple-fsdp,
      title={SimpleFSDP: Simpler Fully Sharded Data Parallel with torch.compile}, 
      author={Ruisi Zhang and Tianyu Liu and Will Feng and Andrew Gu and Sanket Purandare and Wanchao Liang and Francisco Massa},
      year={2024},
      eprint={2411.00284},
      archivePrefix={arXiv},
      primaryClass={cs.DC},
      url={https://arxiv.org/abs/2411.00284}, 
}

@misc{deep-compile,
      title={DeepCompile: A Compiler-Driven Approach to Optimizing Distributed Deep Learning Training}, 
      author={Masahiro Tanaka and Du Li and Umesh Chand and Ali Zafar and Haiying Shen and Olatunji Ruwase},
      year={2025},
      eprint={2504.09983},
      archivePrefix={arXiv},
      primaryClass={cs.DC},
      url={https://arxiv.org/abs/2504.09983}, 
}

@misc{gspmd,
      title={GSPMD: General and Scalable Parallelization for ML Computation Graphs}, 
      author={Yuanzhong Xu and HyoukJoong Lee and Dehao Chen and Blake Hechtman and Yanping Huang and Rahul Joshi and Maxim Krikun and Dmitry Lepikhin and Andy Ly and Marcello Maggioni and Ruoming Pang and Noam Shazeer and Shibo Wang and Tao Wang and Yonghui Wu and Zhifeng Chen},
      year={2021},
      eprint={2105.04663},
      archivePrefix={arXiv},
      primaryClass={cs.DC},
      url={https://arxiv.org/abs/2105.04663}, 
}

@inproceedings{pytorch-2,
author = {Ansel, Jason and Yang, Edward and He, Horace and Gimelshein, Natalia and Jain, Animesh and Voznesensky, Michael and Bao, Bin and Bell, Peter and Berard, David and Burovski, Evgeni and Chauhan, Geeta and Chourdia, Anjali and Constable, Will and Desmaison, Alban and DeVito, Zachary and Ellison, Elias and Feng, Will and Gong, Jiong and Gschwind, Michael and Hirsh, Brian and Huang, Sherlock and Kalambarkar, Kshiteej and Kirsch, Laurent and Lazos, Michael and Lezcano, Mario and Liang, Yanbo and Liang, Jason and Lu, Yinghai and Luk, C. K. and Maher, Bert and Pan, Yunjie and Puhrsch, Christian and Reso, Matthias and Saroufim, Mark and Siraichi, Marcos Yukio and Suk, Helen and Zhang, Shunting and Suo, Michael and Tillet, Phil and Zhao, Xu and Wang, Eikan and Zhou, Keren and Zou, Richard and Wang, Xiaodong and Mathews, Ajit and Wen, William and Chanan, Gregory and Wu, Peng and Chintala, Soumith},
title = {PyTorch 2: Faster Machine Learning Through Dynamic Python Bytecode Transformation and Graph Compilation},
year = {2024},
isbn = {9798400703850},
publisher = {Association for Computing Machinery},
address = {New York, NY, USA},
url = {https://doi.org/10.1145/3620665.3640366},
doi = {10.1145/3620665.3640366},
booktitle = {Proceedings of the 29th ACM International Conference on Architectural Support for Programming Languages and Operating Systems, Volume 2},
pages = {929–947},
numpages = {19},
location = {La Jolla, CA, USA},
series = {ASPLOS '24}
}

@misc{pytorch-ac,
author = {Chillee},
title = {Min-cut optimal recomputation (i.e. activation checkpointing) with AOTAutograd},
date = {2022-01-10},
urldate = {2025-09-20}
}

@misc{gpipe,
      title={GPipe: Efficient Training of Giant Neural Networks using Pipeline Parallelism}, 
      author={Yanping Huang and Youlong Cheng and Ankur Bapna and Orhan Firat and Mia Xu Chen and Dehao Chen and HyoukJoong Lee and Jiquan Ngiam and Quoc V. Le and Yonghui Wu and Zhifeng Chen},
      year={2019},
      eprint={1811.06965},
      archivePrefix={arXiv},
      primaryClass={cs.CV},
      url={https://arxiv.org/abs/1811.06965}, 
}

@misc{gshard,
      title={GShard: Scaling Giant Models with Conditional Computation and Automatic Sharding}, 
      author={Dmitry Lepikhin and HyoukJoong Lee and Yuanzhong Xu and Dehao Chen and Orhan Firat and Yanping Huang and Maxim Krikun and Noam Shazeer and Zhifeng Chen},
      year={2020},
      eprint={2006.16668},
      archivePrefix={arXiv},
      primaryClass={cs.CL},
      url={https://arxiv.org/abs/2006.16668}, 
}

@misc{deepseek-v3,
      title={DeepSeek-V3 Technical Report}, 
      author={DeepSeek-AI and Aixin Liu and Bei Feng and Bing Xue and Bingxuan Wang and Bochao Wu and Chengda Lu and Chenggang Zhao and Chengqi Deng and Chenyu Zhang and Chong Ruan and Damai Dai and Daya Guo and Dejian Yang and Deli Chen and Dongjie Ji and Erhang Li and Fangyun Lin and Fucong Dai and Fuli Luo and Guangbo Hao and Guanting Chen and Guowei Li and H. Zhang and Han Bao and Hanwei Xu and Haocheng Wang and Haowei Zhang and Honghui Ding and Huajian Xin and Huazuo Gao and Hui Li and Hui Qu and J. L. Cai and Jian Liang and Jianzhong Guo and Jiaqi Ni and Jiashi Li and Jiawei Wang and Jin Chen and Jingchang Chen and Jingyang Yuan and Junjie Qiu and Junlong Li and Junxiao Song and Kai Dong and Kai Hu and Kaige Gao and Kang Guan and Kexin Huang and Kuai Yu and Lean Wang and Lecong Zhang and Lei Xu and Leyi Xia and Liang Zhao and Litong Wang and Liyue Zhang and Meng Li and Miaojun Wang and Mingchuan Zhang and Minghua Zhang and Minghui Tang and Mingming Li and Ning Tian and Panpan Huang and Peiyi Wang and Peng Zhang and Qiancheng Wang and Qihao Zhu and Qinyu Chen and Qiushi Du and R. J. Chen and R. L. Jin and Ruiqi Ge and Ruisong Zhang and Ruizhe Pan and Runji Wang and Runxin Xu and Ruoyu Zhang and Ruyi Chen and S. S. Li and Shanghao Lu and Shangyan Zhou and Shanhuang Chen and Shaoqing Wu and Shengfeng Ye and Shengfeng Ye and Shirong Ma and Shiyu Wang and Shuang Zhou and Shuiping Yu and Shunfeng Zhou and Shuting Pan and T. Wang and Tao Yun and Tian Pei and Tianyu Sun and W. L. Xiao and Wangding Zeng and Wanjia Zhao and Wei An and Wen Liu and Wenfeng Liang and Wenjun Gao and Wenqin Yu and Wentao Zhang and X. Q. Li and Xiangyue Jin and Xianzu Wang and Xiao Bi and Xiaodong Liu and Xiaohan Wang and Xiaojin Shen and Xiaokang Chen and Xiaokang Zhang and Xiaosha Chen and Xiaotao Nie and Xiaowen Sun and Xiaoxiang Wang and Xin Cheng and Xin Liu and Xin Xie and Xingchao Liu and Xingkai Yu and Xinnan Song and Xinxia Shan and Xinyi Zhou and Xinyu Yang and Xinyuan Li and Xuecheng Su and Xuheng Lin and Y. K. Li and Y. Q. Wang and Y. X. Wei and Y. X. Zhu and Yang Zhang and Yanhong Xu and Yanhong Xu and Yanping Huang and Yao Li and Yao Zhao and Yaofeng Sun and Yaohui Li and Yaohui Wang and Yi Yu and Yi Zheng and Yichao Zhang and Yifan Shi and Yiliang Xiong and Ying He and Ying Tang and Yishi Piao and Yisong Wang and Yixuan Tan and Yiyang Ma and Yiyuan Liu and Yongqiang Guo and Yu Wu and Yuan Ou and Yuchen Zhu and Yuduan Wang and Yue Gong and Yuheng Zou and Yujia He and Yukun Zha and Yunfan Xiong and Yunxian Ma and Yuting Yan and Yuxiang Luo and Yuxiang You and Yuxuan Liu and Yuyang Zhou and Z. F. Wu and Z. Z. Ren and Zehui Ren and Zhangli Sha and Zhe Fu and Zhean Xu and Zhen Huang and Zhen Zhang and Zhenda Xie and Zhengyan Zhang and Zhewen Hao and Zhibin Gou and Zhicheng Ma and Zhigang Yan and Zhihong Shao and Zhipeng Xu and Zhiyu Wu and Zhongyu Zhang and Zhuoshu Li and Zihui Gu and Zijia Zhu and Zijun Liu and Zilin Li and Ziwei Xie and Ziyang Song and Ziyi Gao and Zizheng Pan},
      year={2025},
      eprint={2412.19437},
      archivePrefix={arXiv},
      primaryClass={cs.CL},
      url={https://arxiv.org/abs/2412.19437}, 
}

@misc{deepseekmoe,
      title={DeepSeekMoE: Towards Ultimate Expert Specialization in Mixture-of-Experts Language Models}, 
      author={Damai Dai and Chengqi Deng and Chenggang Zhao and R. X. Xu and Huazuo Gao and Deli Chen and Jiashi Li and Wangding Zeng and Xingkai Yu and Y. Wu and Zhenda Xie and Y. K. Li and Panpan Huang and Fuli Luo and Chong Ruan and Zhifang Sui and Wenfeng Liang},
      year={2024},
      eprint={2401.06066},
      archivePrefix={arXiv},
      primaryClass={cs.CL},
      url={https://arxiv.org/abs/2401.06066}, 
}

@inproceedings {tvm,
author = {Tianqi Chen and Thierry Moreau and Ziheng Jiang and Lianmin Zheng and Eddie Yan and Haichen Shen and Meghan Cowan and Leyuan Wang and Yuwei Hu and Luis Ceze and Carlos Guestrin and Arvind Krishnamurthy},
title = {{TVM}: An Automated {End-to-End} Optimizing Compiler for Deep Learning},
booktitle = {13th USENIX Symposium on Operating Systems Design and Implementation (OSDI 18)},
year = {2018},
isbn = {978-1-939133-08-3},
address = {Carlsbad, CA},
pages = {578--594},
url = {https://www.usenix.org/conference/osdi18/presentation/chen},
publisher = {USENIX Association},
month = oct
}

@inproceedings {mirage,
title={Mirage: A Multi-Level Superoptimizer for Tensor Programs}, 
author={Mengdi Wu and Xinhao Cheng and Shengyu Liu and Chunan Shi and Jianan Ji and Kit Ao and Praveen Velliengiri and Xupeng Miao and Oded Padon and Zhihao Jia},
booktitle = {19th USENIX Symposium on Operating Systems Design and Implementation (OSDI 25)},
year = {2025},
address = {Boston, MA},
publisher = {USENIX Association},
month = jul
}

@software{aitemplate,
  author       = {{Meta}},
  title        = {{AITemplate: Python framework which renders neural network into high performance CUDA/HIP C++ code.}},
  publisher    = {facebookincubator},
  year  = {2022},
  version      = {main},
  url          = {https://github.com/facebookincubator/AITemplate},
}

@misc{checkmate,
      title={Checkmate: Breaking the Memory Wall with Optimal Tensor Rematerialization}, 
      author={Paras Jain and Ajay Jain and Aniruddha Nrusimha and Amir Gholami and Pieter Abbeel and Kurt Keutzer and Ion Stoica and Joseph E. Gonzalez},
      year={2020},
      eprint={1910.02653},
      archivePrefix={arXiv},
      primaryClass={cs.LG},
      url={https://arxiv.org/abs/1910.02653}, 
}

@misc{dynamic-tensor-remat,
      title={Dynamic Tensor Rematerialization}, 
      author={Marisa Kirisame and Steven Lyubomirsky and Altan Haan and Jennifer Brennan and Mike He and Jared Roesch and Tianqi Chen and Zachary Tatlock},
      year={2021},
      eprint={2006.09616},
      archivePrefix={arXiv},
      primaryClass={cs.LG},
      url={https://arxiv.org/abs/2006.09616}, 
}

@misc{sublinear-memory-cost,
      title={Training Deep Nets with Sublinear Memory Cost}, 
      author={Tianqi Chen and Bing Xu and Chiyuan Zhang and Carlos Guestrin},
      year={2016},
      eprint={1604.06174},
      archivePrefix={arXiv},
      primaryClass={cs.LG},
      url={https://arxiv.org/abs/1604.06174}, 
}

@misc{simple-document-understanding,
      title={A Simple yet Effective Layout Token in Large Language Models for Document Understanding}, 
      author={Zhaoqing Zhu and Chuwei Luo and Zirui Shao and Feiyu Gao and Hangdi Xing and Qi Zheng and Ji Zhang},
      year={2025},
      eprint={2503.18434},
      archivePrefix={arXiv},
      primaryClass={cs.CV},
      url={https://arxiv.org/abs/2503.18434}, 
}

@misc{mdoc,
      title={MDocAgent: A Multi-Modal Multi-Agent Framework for Document Understanding}, 
      author={Siwei Han and Peng Xia and Ruiyi Zhang and Tong Sun and Yun Li and Hongtu Zhu and Huaxiu Yao},
      year={2025},
      eprint={2503.13964},
      archivePrefix={arXiv},
      primaryClass={cs.LG},
      url={https://arxiv.org/abs/2503.13964}, 
}

@misc{dude,
      title={Document Understanding Dataset and Evaluation (DUDE)}, 
      author={Jordy Van Landeghem and Rubén Tito and Łukasz Borchmann and Michał Pietruszka and Paweł Józiak and Rafał Powalski and Dawid Jurkiewicz and Mickaël Coustaty and Bertrand Ackaert and Ernest Valveny and Matthew Blaschko and Sien Moens and Tomasz Stanisławek},
      year={2023},
      eprint={2305.08455},
      archivePrefix={arXiv},
      primaryClass={cs.CV},
      url={https://arxiv.org/abs/2305.08455}, 
}

@misc{gsm8k,
      title={Training Verifiers to Solve Math Word Problems}, 
      author={Karl Cobbe and Vineet Kosaraju and Mohammad Bavarian and Mark Chen and Heewoo Jun and Lukasz Kaiser and Matthias Plappert and Jerry Tworek and Jacob Hilton and Reiichiro Nakano and Christopher Hesse and John Schulman},
      year={2021},
      eprint={2110.14168},
      archivePrefix={arXiv},
      primaryClass={cs.LG},
      url={https://arxiv.org/abs/2110.14168}, 
}

@misc{smore,
      title={Scalable Multi-Hop Relational Reasoning for Knowledge-Aware Question Answering}, 
      author={Yanlin Feng and Xinyue Chen and Bill Yuchen Lin and Peifeng Wang and Jun Yan and Xiang Ren},
      year={2020},
      eprint={2005.00646},
      archivePrefix={arXiv},
      primaryClass={cs.CL},
      url={https://arxiv.org/abs/2005.00646}, 
}

@inproceedings{mt-bench,
   title={MT-Bench-101: A Fine-Grained Benchmark for Evaluating Large Language Models in Multi-Turn Dialogues},
   url={http://dx.doi.org/10.18653/v1/2024.acl-long.401},
   DOI={10.18653/v1/2024.acl-long.401},
   booktitle={Proceedings of the 62nd Annual Meeting of the Association for Computational Linguistics (Volume 1: Long Papers)},
   publisher={Association for Computational Linguistics},
   author={Bai, Ge and Liu, Jie and Bu, Xingyuan and He, Yancheng and Liu, Jiaheng and Zhou, Zhanhui and Lin, Zhuoran and Su, Wenbo and Ge, Tiezheng and Zheng, Bo and Ouyang, Wanli},
   year={2024},
   pages={7421–7454} }

@misc{llama-2,
      title={Llama 2: Open Foundation and Fine-Tuned Chat Models}, 
      author={Hugo Touvron and Louis Martin and Kevin Stone and Peter Albert and Amjad Almahairi and Yasmine Babaei and Nikolay Bashlykov and Soumya Batra and Prajjwal Bhargava and Shruti Bhosale and Dan Bikel and Lukas Blecher and Cristian Canton Ferrer and Moya Chen and Guillem Cucurull and David Esiobu and Jude Fernandes and Jeremy Fu and Wenyin Fu and Brian Fuller and Cynthia Gao and Vedanuj Goswami and Naman Goyal and Anthony Hartshorn and Saghar Hosseini and Rui Hou and Hakan Inan and Marcin Kardas and Viktor Kerkez and Madian Khabsa and Isabel Kloumann and Artem Korenev and Punit Singh Koura and Marie-Anne Lachaux and Thibaut Lavril and Jenya Lee and Diana Liskovich and Yinghai Lu and Yuning Mao and Xavier Martinet and Todor Mihaylov and Pushkar Mishra and Igor Molybog and Yixin Nie and Andrew Poulton and Jeremy Reizenstein and Rashi Rungta and Kalyan Saladi and Alan Schelten and Ruan Silva and Eric Michael Smith and Ranjan Subramanian and Xiaoqing Ellen Tan and Binh Tang and Ross Taylor and Adina Williams and Jian Xiang Kuan and Puxin Xu and Zheng Yan and Iliyan Zarov and Yuchen Zhang and Angela Fan and Melanie Kambadur and Sharan Narang and Aurelien Rodriguez and Robert Stojnic and Sergey Edunov and Thomas Scialom},
      year={2023},
      eprint={2307.09288},
      archivePrefix={arXiv},
      primaryClass={cs.CL},
      url={https://arxiv.org/abs/2307.09288}, 
}
\bibliographystyle{iclr2026_conference}


\end{document}